\documentclass[10pt,twocolumn,a4paper]{esaAI}
\usepackage[dvipsnames]{xcolor}

\title{EclipseNETs: a differentiable description of irregular eclipse conditions.}

\def\authorEmail{dario.izzo@esa.int}
\def\authoremail{g.acciarini@surrey.ac.uk}

\author[1,2]{Giacomo Acciarini\thanks{Corresponding author. \authoremail}}
\author[2]{Francesco Biscani}
\author[1]{Dario Izzo\thanks{Corresponding author. \authorEmail}}

\affil[1]{Advanced Concepts Team, European Space Research and Technology Centre (ESTEC), Noordwijk, The Netherlands.}
\affil[2]{Surrey Space Centre, University of Surrey, Guildford, United Kingdom.}
\affil[3]{ESA/ESOC, Darmstadt, Germany}

\begin{document}

\makeCustomtitle

\begin{abstract}
In the field of spaceflight mechanics and astrodynamics, determining eclipse regions is a frequent and critical challenge. This determination impacts various factors, including the acceleration induced by solar radiation pressure, the spacecraft power input, and its thermal state all of which must be accounted for in various phases of the mission design. This study leverages recent advances in neural image processing to develop fully differentiable models of eclipse regions for highly irregular celestial bodies. By utilizing test cases involving Solar System bodies previously visited by spacecraft, such as 433 Eros, 25143 Itokawa, 67P/Churyumov–Gerasimenko, and 101955 Bennu, we propose and study an implicit neural architecture defining the shape of the eclipse cone based on the Sun's direction. Employing periodic activation functions, we achieve high precision in modeling eclipse conditions. Furthermore, we discuss the potential applications of these differentiable models in spaceflight mechanics computations.
\end{abstract}

\section{Introduction}

Estimating the geometry of eclipses during satellite motion is often important, as they influence the solar radiation pressure-induced acceleration, as well as communication, thermal, and power system design \cite{adhya2004oblate, jia2017eclipse, srivastava2015eclipse}.
When an object with a high area-to-mass ratio is in motion around small bodies with a weak gravitational potential, the solar radiation pressure can become one of the main perturbing forces, and the precise computation of eclipse conditions becomes essential to have an accurate prediction of their motion and its stability \cite{lang2022heliotropic, lang2021spacecraft, mcmahon2020dynamical}. The eclipse computation involves the analysis of whether the time-dependent state vector describing the orbiting object's position lies in the shadow cast by the celestial body \cite{neta1998satellite}. Different cases may arise according to the shape and dimension of the shadowing body. The eclipse cone may be approximated to a cylinder and the body shape as a sphere or an ellipsoid. A full 3D model of the body shape, when available may also be necessary to meet highly precise predictions. In the case of the small irregular body 101955 Bennu, visited by NASA OSIRIS-REx spacecraft in 2019~\cite{malyuta2021advances}, orbital simulations of ejected pebbles found orbiting the asteroid \cite{hergenrother2019operational} required precise modelling of its shadowing effects \cite{mcmahon2020dynamical} and used the highly precise Bennu's 3D model. Assuming a cylindrical shadow area, exact eclipse conditions can be then carried out using Möller–Trumbore intersection algorithm \cite{moller2005fast}, or any equivalent ray-tracing approach, resulting in a computation that cannot thus be used to determine the entrance and exit conditions of the eclipse reliably (see \cite{biscani2022reliable}) nor to efficiently build propagators \cite{mcmahon2020dynamical}. 

In this paper, we introduce EclipseNETs, a technique to obtain a neural-based implicit representation of cylindrical shadows cast by irregular bodies. This novel approach leverages the latest advances in neural processing of images to model complex geometries in a fully differentiable and computationally efficient manner, enabling highly precise numerical propagation of orbits that include solar radiation pressure effects. Our method builds on recent advances in implicit neural representations, particularly Neural Radiance Fields (NeRFs) and their variants, which have demonstrated remarkable success in capturing intricate details of 3D scenes from 2D images through neural networks trained on photometric consistency \cite{mildenhall2021nerf}.
Implicit neural representations, as explored in the works of Sitzmann et al. \cite{sitzmann2020implicit}, utilize periodic activation functions to represent high-frequency details and complex shapes implicitly. These implicit representations are particularly well-suited for modeling the shadows cast by celestial bodies because they can accurately and efficiently represent the intricate geometries and dynamics involved.

Our approach integrates these ideas into the domain of orbital dynamics, where traditional methods like ray-tracing are limited by their lack of differentiability and computational inefficiencies. By using neural implicit representations, EclipseNETs can handle the precise calculation of eclipse conditions more robustly. This technique transforms the problem into one that can be efficiently solved with Taylor-based methods, ultimately improving the reliability and accuracy of eclipse event detection during orbital propagation.

\section{Results}
\label{sec:results}
We represent eclipse conditions using an implicit neural representation called EclipseNET. EclipseNET is a neural architecture with periodic activation functions \cite{sitzmann2020implicit}, capable of capturing the complex geometrical details of eclipses cast by irregular bodies in any direction. The only inputs to EclipseNET are the Cartesian coordinates of the spacecraft's position and the Sun's direction. It outputs the eclipse function $F$ introduced in this work, which is a scalar value: negative when in eclipse, positive otherwise. 
We significantly improve and extend a preliminary study of a related architecture \cite{biscani2022reliable} by redefining the eclipse function and using periodic activations, resulting in substantial improvements in the implicit representations obtained. Additionally, we develop the proposed representation for several widely studied irregular bodies. Figure \ref{fig:plot_small_bodies} D) illustrates how, after training, our EclipseNETs accurately capture the eclipse condition $F=0$. Although the figure shows results for only one specific Sun direction, the overall performance of the networks (in terms of the loss achieved) is reported in Table \ref{tab:ml_models_results}, confirming the high level of approximation achieved in all Sun directions and the superiority of the SIREN architecture over a more standard ReLU-based pipeline.

\begin{figure*}[tb]
  \centering
  \includegraphics[width=2.\columnwidth]{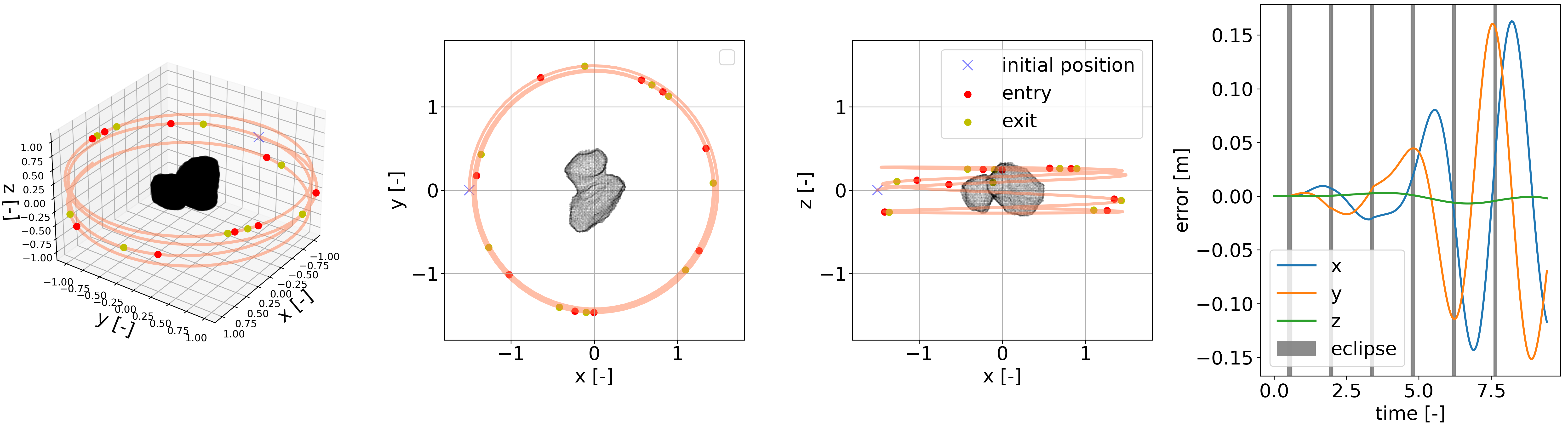}
  \caption{On the left three panels: three views of a spacecraft trajectory around Churyumov-Gerasimenko, with initial conditions, and entry and exit eclipse conditions highlighted; on the right panel: error in positional coordinates between a trajectory found computing the silhouette with Möller–Trumbore, against the one found using a neural network.}
	\label{fig:plot_orbits_w_eclipses}
\end{figure*}

Once an EclipseNET is trained and achieves a low loss, there may still be a minor mismatch between the actual eclipse cylinder and the one implicitly represented by the neural model (as visualized in Figure \ref{fig:plot_small_bodies} D)). Figure \ref{fig:plot_orbits_w_eclipses}, demonstrates the error introduced when simulating a few orbits that traverse several eclipses around the highly irregular body of the comet 67P/Churyumov–Gerasimenko (the most complex out of our four cases) with an EclipseNET of 2,369 learnable parameters. After a few orbits, the error remains within centimeters, while the simulation speed improves significantly as the exact ray-tracing  Möller–Trumbore \cite{moller2005fast} triangle intersection algorithm (vectorized) employed to compute the ground truth reveals to be more than two orders of magnitude slower than the network inference. Similar results were obtained with different initial conditions and body shapes, suggesting that EclipseNETs can effectively replace, in the context of space-flight mechanics simulations, complex ray-tracing algorithms.

\section{Methods}
\subsection{Eclipse Function}
\label{sec:eclipse_function}
We represent implicitly the eclipsed area originated by an irregular body $\mathcal B$ when illuminated by a point source placed at infinite in the direction $\hat{\mathbf s}$ (body frame) by introducing an Eclipse function $F_{\mathcal B}(x,y, \hat{\mathbf s})$. In a reference frame defined in a plane orthogonal to $\hat{\mathbf s}$, the eclipse function is defined as:
$$
F_{\mathcal B}(x,y, \hat{\mathbf s}) = \left\{
\begin{array}{l}
d(x,y, \partial \Omega_{\hat{\mathbf s}})\\
-d(x,y, \partial \Omega_{\hat{\mathbf s}})\\
\end{array}
\right.
$$
where $d$ is the distance of the point located in $x,y$ to the border $\partial \Omega_{\hat{\mathbf s}}$ of the eclipse zone. The concept is similar to the signed distance function, widely used in computer graphics \cite{park2019deepsdf}, but it represents two-dimensional shapes parameterized by a projection direction ${\hat{\mathbf s}}$. Examples of the eclipse function for a specific direction are shown in Figure~\ref{fig:plot_small_bodies} B) for all bodies. Note that we are here using a different definition of the Eclipse Function than the one that appeared in our previous work \cite{biscani2022reliable}. The advantage of the new definition stems from the increased smoothness of the resulting function and from being now able to write the differential property that captures the gradient of the eclipse function in the direction perpendicular to the eclipse area as:
$$
\nabla_{x,y} F_{\mathcal B} \cdot \hat{\mathbf n}= 1.
$$
where $\hat{\mathbf n}$ is the normal (outward or inward) to $\partial \Omega_{\hat{\mathbf s}}$\begin{figure}[tb]
  \centering
  \includegraphics[width=0.8\columnwidth]{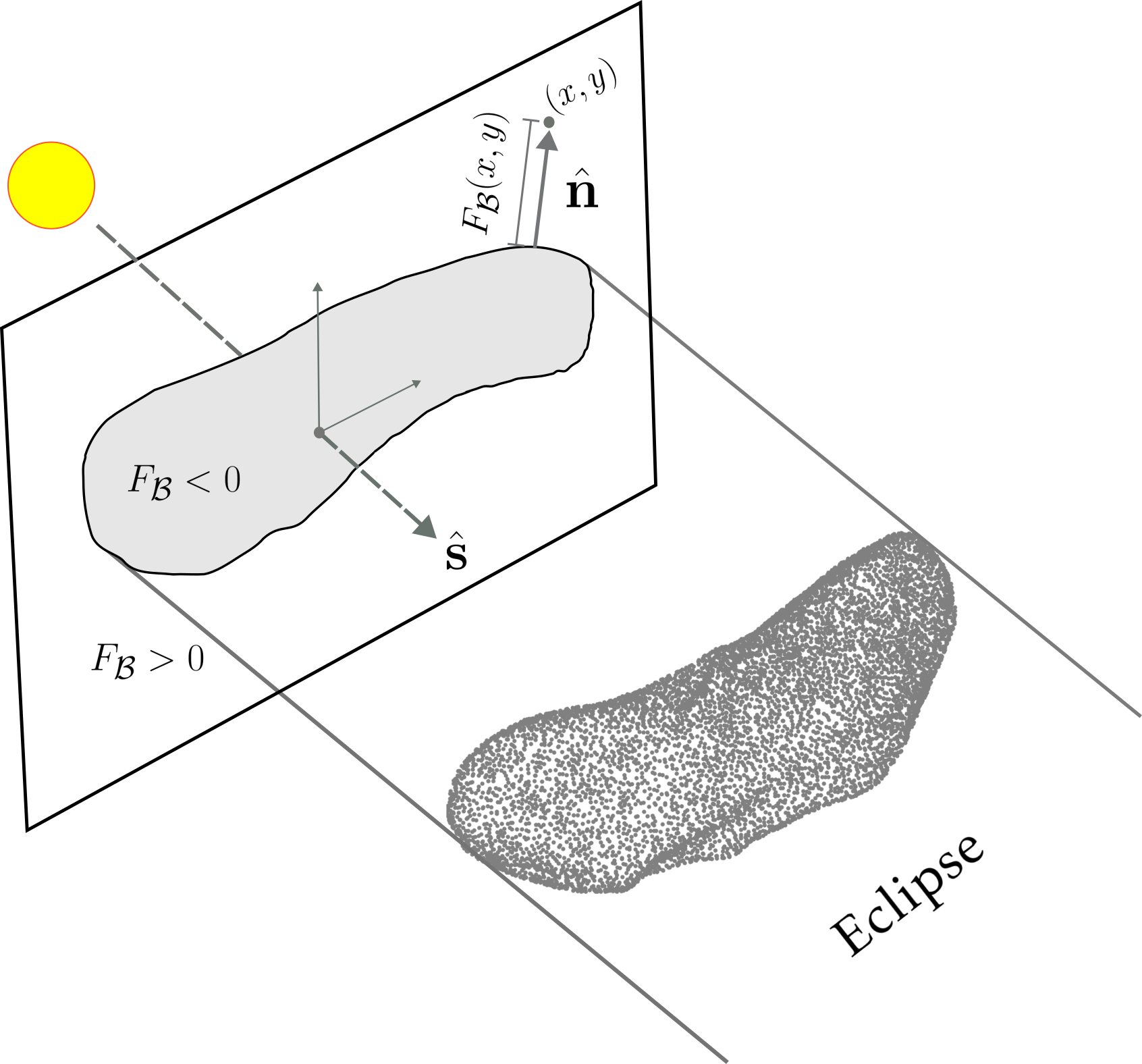}
  \caption{Basic definition of the eclipse function $F_\mathcal B$ here introduced and other relevant quantities.}
	\label{fig:eclipse_geometry}
\end{figure}

\subsection{Irregular Bodies}
We primarily focus on four different small bodies: 101955 Bennu, 67/P Churyumov-Gerasimenko, 433 Eros, and 25143 Itokawa. 
In all cases, we utilize high-fidelity polyhedral models reconstructed from various instruments on board spacecraft that visited these celestial bodies. For Eros and Itokawa, we use models produced by Robert Gaskell \cite{erospoly} \cite{itokawapoly}; for Bennu, we employ the model provided by the OSIRIS-REx team \cite{bennupoly}; and for Churyumov–Gerasimenko, we use the model available from the European Space Agency \cite{67ppoly}. For the purpose of modelling the gravitational field of these bodies, we transform the surface meshes into mascon models. This is done by first creating a constrained Delaunay tetrahedralization \cite{si2015tetgen} and then placing a mass at the centroid of each resulting tetrahedron. In igure~\ref{fig:plot_small_bodies} A), we display a 3D view of each of these models. Eventually, all asteroid models consist of a triangular mesh, describing the asteroid surface, and a mascon model describing the asteroid mass distribution here considered as uniform in all cases \cite{izzo2022geodesy}. It is clear that the number of mesh points, and mascons, as well as the rotation period of the body and its mass, will influence the dynamics of an orbiting spacecraft. In Table~\ref{tab:parameters_asteroids_comets}, we report these values for all four bodies, together with the characteristic length used to normalize all positional coordinates of the shape models and spacecraft state.

\begin{table}[h!]
\scriptsize
    \centering
    \begin{tabular}{|l|c|c|c|c|c|c|}
\hline
        \ & N mesh & N mascons & $\omega$ [hr] & L [km] & mass [kg] \\
\hline
        Bennu                   &  7,374                  & 75,150                 & 4.296        &  0.5634   & $7.329\times 10^{10}$ \\
        Itokawa                 & 3000                   & 100,363                 & 12.132        &  0.5607    &  $3.51\times 10^{10}$ \\
        67/P                 & 9,149                   & 57,259                 & 12.4043             &  5.0025   &  $9.982\times10^{12}$  \\
        Eros                    & 7,374                   & 97,824                 & 5.270              & 32.6622   & $6.687 \times 10^{15}$  \\
\hline
    \end{tabular}
    \caption{Data on the 3D models used.}
    \label{tab:parameters_asteroids_comets}
\end{table}

The 3D triangular mesh is used to run the Möller–Trumbore intersection algorithm \cite{moller2005fast} to establish the ground truth of a point being eclipsed or not. The mascon model is used in the trajectory simulations to represent the irregular gravity field of the asteroid.

\subsection{Spacecraft dynamics}
We describe the spacecraft dynamics in a body-centered reference frame using the following set of differential equations:
\begin{equation}
    \label{eq:eom_mascon}
\ddot {\mathbf r} = -G \sum_{j=0}^N \frac {m_j}{|\mathbf r - \mathbf r_j|^3} (\mathbf r - \mathbf r_j) - 2 \boldsymbol\omega \times \mathbf v - \boldsymbol \omega \times\boldsymbol\omega \times \mathbf r - \eta \hat{\mathbf s}(t),
\end{equation}
where $\mathbf r$ is the spacecraft position, $m_j$ represent the various mascon masses describing the asteroid irregular field, $\eta$ is the acceleration due to the solar radiation pressure (zero when in eclipse) and $\boldsymbol \omega$ the body angular velocity.
Note that the Sun direction $\hat{\mathbf s}(t) = \mathbf R(t)\hat{\mathbf s}(0)$ is modeled, neglecting the asteroid orbital motion, rotating the initial Sun direction $\hat{\mathbf s}(0)$ around the asteroid rotation axis, of an angle $\omega t$. 

\subsection{Training and dataset creation}

\begin{figure*}[tb]
  \centering
  \includegraphics[width=2.\columnwidth]{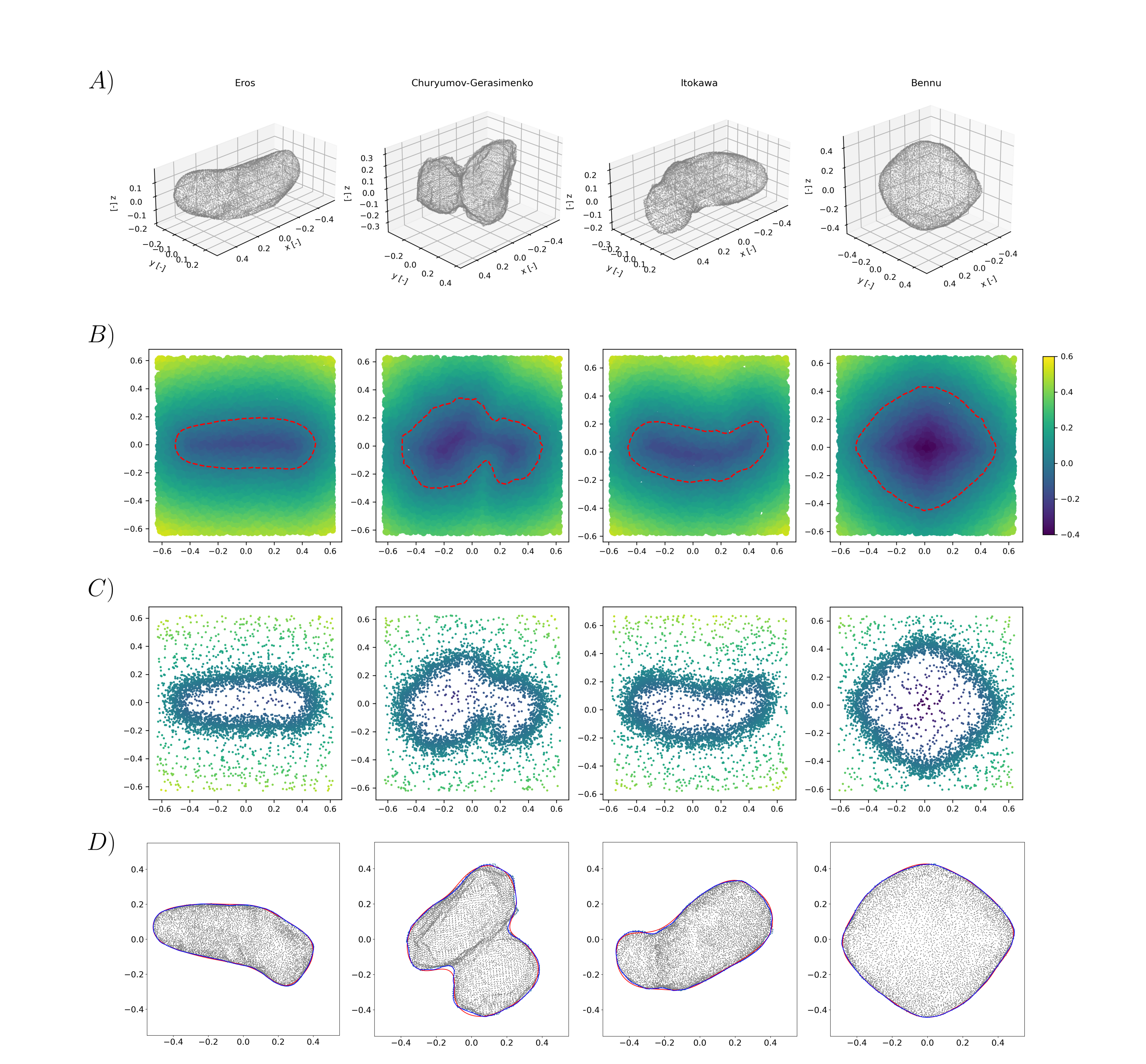}
  \caption{A) 3D model for of Bennu, Churyumov-Gerasimenko, Eros, and Itokawa. B): contour plot of the eclipse function, for a fixed view; C): examples of points where the eclipse function was sampled to construct the training set. D) Predictions of the eclipse for a Sun direction not on the training set. In red, an EclipseNet of 2,369 was used, in blue 50,561 parameters.}
	\label{fig:plot_small_bodies}
\end{figure*}

To create a dataset containing ground truth values for the eclipse function $F(x,y,\hat{\mathbf{s}})$, we consider an isotropic range of Sun directions $\hat{\mathbf{s}}$. We employ the Fibonacci sphere and generate 500 directions for training and 200 for validation. For each, we uniformly sample 1,000 points $(x,y)$ in the $[-1,1]$ range (all models are considered as non dimensional, see Table \ref{tab:parameters_asteroids_comets}), and 3,000 points within a small radius around the asteroid eclipse border $\partial \Omega_{\hat{\mathbf{s}}}$. This process yields a distribution of points illustrated, for example, in Figure~\ref{fig:plot_small_bodies} C) for a particular Sun direction. A total of 22 million points for training and 9 million points for validation are generated for each of the four small bodies. On these dataset, SIREN networks having 3 hidden layers and 32 neurons per layer are trained. We use the training pipeline suggested in \cite{sitzmann2020implicit} taking care accordingly of weight initialization and input normalization.

For comparison we also train a more standard model using ReLU nonlinearities having the same size of learnable parameters: 2,369. In Table~\ref{tab:ml_models_results} we display the obtained losses, which highlights that both in training and validation, Siren networks outperform ReLU-based architectures. 

Both ReLU and Siren networks are trained with a minibatch size of 256, an initial learning rate parameter of $3\times10^{-4}$, a total of 60 epochs, and a multi-step learning rate scheduler that reduces the learning rate with a decay factor of 0.7, once the number of epoch reaches 25, and every next five epochs.

Finally, to qualitatively show the improvement obtained by employing larger networks, in Figure~\ref{fig:plot_small_bodies} D), we display an example of inference from the validation set using the 2,369 parameters siren network, and one with about 20 times more parameters in blue (i.e., 50,561). As it can be seen, the bigger networks can reconstruct the asteroid silhouette more accurately. 

\begin{table}[h!]
\scriptsize
    \centering
    \begin{tabular}{|l|c|c|c|c|c|c|}
\hline
        \ &  $\mathcal{L}_{MSE}$ ReLU &  $\mathcal{L}_{MSE}$ Siren &  Dataset\\
\hline
         $\mathcal N_{Bennu}$ & $3.5295\times 10^{-5}$ &$1.99059\times 10^{-5}$ & $\mathcal{D}_{train,Bennu}$\\
         $\mathcal{N}_{Itokawa}$  &  $8.0184\times10^{-5}$ &  $4.7082\times 10^{-5}$ & $\mathcal{D}_{train,Itokawa}$\\
       $\mathcal{N}_{67/P}$ & $1.49711\times 10^{-4}$  &    $7.3122\times 10^{-5}$ &  $\mathcal{D}_{train,67/P}$\\
        $\mathcal{N}_{Eros}$ &  $9.3738\times 10^{-5}$ &   $4.2137\times 10^{-5}$ &  $\mathcal{D}_{train,Eros}$\\
\hline
         $\mathcal N_{Bennu}$ &  $3.6508\times 10^{-5}$ & $2.1773\times 10^{-5}$ & $\mathcal{D}_{valid,Bennu}$\\
         $\mathcal{N}_{Itokawa}$  & $8.4136\times 10^{-5}$ & $4.9863\times 10^{-5}$ & $\mathcal{D}_{valid,Itokawa}$\\
       $\mathcal{N}_{67/P}$ & $1.59457\times 10^{-4}$ &  $7.75078\times 10^{-5}$ &  $\mathcal{D}_{valid,67/P}$\\
        $\mathcal{N}_{Eros}$ & $9.8826\times 10^{-5}$ & $4.5850\times 10^{-5}$ &  $\mathcal{D}_{valid,Eros}$\\
    \hline
    \end{tabular}
    \caption{Feed forward neural networks with ReLU activation function and Siren network models mean squared error loss (MSE) on the training and validation datasets for all four small bodies.}
    \label{tab:ml_models_results}
\end{table}

\section{Author Contributions}
D.I. and F.B. conceived the project and made a preliminary analysis and definition of the Eclipse function and its use in space-flight mechanics. G.A. modified and simplified the Eclipse function definition, made the numerical simulations, and trained all EclipseNETs for this work. D.I. made all 3D models for the irregular bodies. G.A, D.I., and F.B. analyzed the results and wrote the paper.

\printbibliography
\addcontentsline{toc}{section}{References}

\end{document}